\documentclass{article}
\usepackage{spconf,amsmath,amssymb,graphicx,hyperref,booktabs}
\hypersetup{hidelinks}
\let\originalthebibliography\thebibliography
\renewcommand{\thebibliography}[1]{\originalthebibliography{#1}\setlength{\itemsep}{0pt}}

\newcommand{\framework}{HiGRTrack}

\title{HiGRTrack: Hierarchical Gated Residual Collaboration for RGB-D-T Object Tracking}

\name{{\fontsize{9.5}{12}\selectfont
Mingzhe Liu$^{1,7}$\hspace{0.3em}Junkai Ding$^{1}$\hspace{0.3em}Zhihao Chen$^{2}$\hspace{0.3em}Yihan Ba$^{1}$\hspace{0.3em}%
Yuk-Kit Hou$^{3,7}$\hspace{0.3em}Junhan Sun$^{4,7}$\hspace{0.3em}Weiyang Huang$^{5}$\hspace{0.3em}Hang Guo$^{6}$\hspace{0.3em}Tao Dai$^{1,6}$}}
\address{{\fontsize{10}{11}\selectfont\shortstack{
$^{1}$ Shenzhen University; \quad $^{2}$ Jilin University; \quad $^{3}$ Peking University; \quad $^{4}$ Zhejiang University\\
$^{5}$ University of Washington; \quad $^{6}$ Tsinghua University; \quad $^{7}$ RoboParty Lab}}}

\begin{document}
\ninept
\maketitle

\begin{abstract}
RGB-D-T tracking combines appearance, geometry, and thermal contrast,
providing complementary evidence for target localization under diverse
conditions. However, the three modalities play different roles, and their
useful information varies across targets, regions, and frames. Direct fusion
can weaken reliable RGB representations and mix redundant responses from depth
and thermal inputs. We propose HiGRTrack (Hierarchical Gated Residual Tracking), a hierarchical collaboration framework
that coordinates the three modalities according to their roles. First, frozen
DINOv3 features enrich RGB semantics, and a residual gate controls their
contribution to the tracker tokens. In parallel, a bidirectional residual interaction processes depth and thermal
features, while spatial and channel routing selects their contributions. The calibrated representations are then
delivered to a frozen tracking backbone through recurrent deep prompts.
Experiments on RGBDT500 show that HiGRTrack achieves 0.759 AUC and 0.804 DP,
and ablation results support the complete configuration.
\end{abstract}


\section{Introduction}

Visual object tracking is an important component of intelligent vision systems,
supporting persistent localization and understanding of dynamic objects in
applications such as autonomous driving, video surveillance, and robotic
perception~\cite{chen2022visual}. Despite advances in Transformer based RGB trackers
~\cite{chen2021transt,yan2021stark,ye2022one,chen2023seqtrack,dosovitskiy2020vit},
visible imagery remains fragile under low illumination, occlusion, and
confusing backgrounds. Depth provides geometric cues for target separation,
whereas thermal imagery provides contrast that is less affected by visible
light. The complementary value of RGB with depth or thermal imagery has
motivated RGB-D datasets such as DepthTrack and
RGBD1K~\cite{yan2021depthtrack,zhu2023rgbd1k}, and RGB-T benchmarks such as
LasHeR~\cite{li2021lasher}. Recent work on RGB-D-T tracking brings
appearance, geometry, and thermal contrast into a common tracking framework,
providing a richer basis for target localization and establishing a recent
direction in multimodal tracking.

Existing multimodal trackers mainly study RGB-D or RGB-T settings and
introduce auxiliary information through feature interaction and prompt
adaptation~\cite{zhu2023vipt,wu2024untrack,hou2024sdstrack,cao2024bat}.
For RGB-D-T tracking, RGBDT500 introduces aligned streams and RDTTrack,
which combines bidirectional depth--thermal residuals, foveated prompts,
and a frozen OSTrack backbone~\cite{zhu2025rgbdt500}. Its pre-prompt
collaboration has two limitations. First, the initial RGB representation
comes solely from the tracking backbone, without a separate feature-enrichment
branch. Second, depth and thermal residuals are concatenated and fused
through a shared $1\times1$ projection. Although the subsequent prompts
apply spatial weighting, this residual-fusion step lacks an explicit
input-conditioned coefficient to balance the two modalities at each position
and channel. Residual extraction therefore leaves a separate selection
problem: deciding how much of each auxiliary residual to retain for the
current input.

HiGRTrack addresses these limitations by adding two feature-dependent gates
while retaining RDTTrack's residual construction and recurrent prompts.
The RGB gate interpolates tracker tokens with DINOv3 features at each token
and channel; the enhanced representation supplies both the backbone and
prompt interface. In parallel, the depth--thermal gate predicts mixing
coefficients from the two residuals before output projection. These changes
enable explicit adjustment of RGB feature enrichment and auxiliary modality
contributions before prompt generation, complementing the inherited spatial
weighting.

On RGBDT500, HiGRTrack improves AUC/DP from RDTTrack's 0.752/0.792 to
0.759/0.804, absolute gains of 0.007/0.012 (Table~\ref{tab:main}).
The No DINOv3 and depthwise-gate variants achieve 0.736/0.784 and
0.744/0.787 AUC/DP, respectively (Table~\ref{tab:ablation}), with the
full configuration performing best among the reported settings.
Our contributions are:
\begin{itemize}
    \item an extension of RDTTrack that coordinates semantic RGB evidence
    and depth and thermal evidence before prompt fusion;
    \item a semantic residual gate for RGB features and a Depth-TIR
    complementarity gate that routes complementary residuals across spatial
    positions and channels; and
    \item a comparison with 20 baseline configurations on RGBDT500, where the proposed method
    achieves the highest AUC of 0.759 and DP of 0.804 among these configurations.
\end{itemize}

\begin{figure*}[t]
    \centering
    \includegraphics[width=0.91\textwidth]{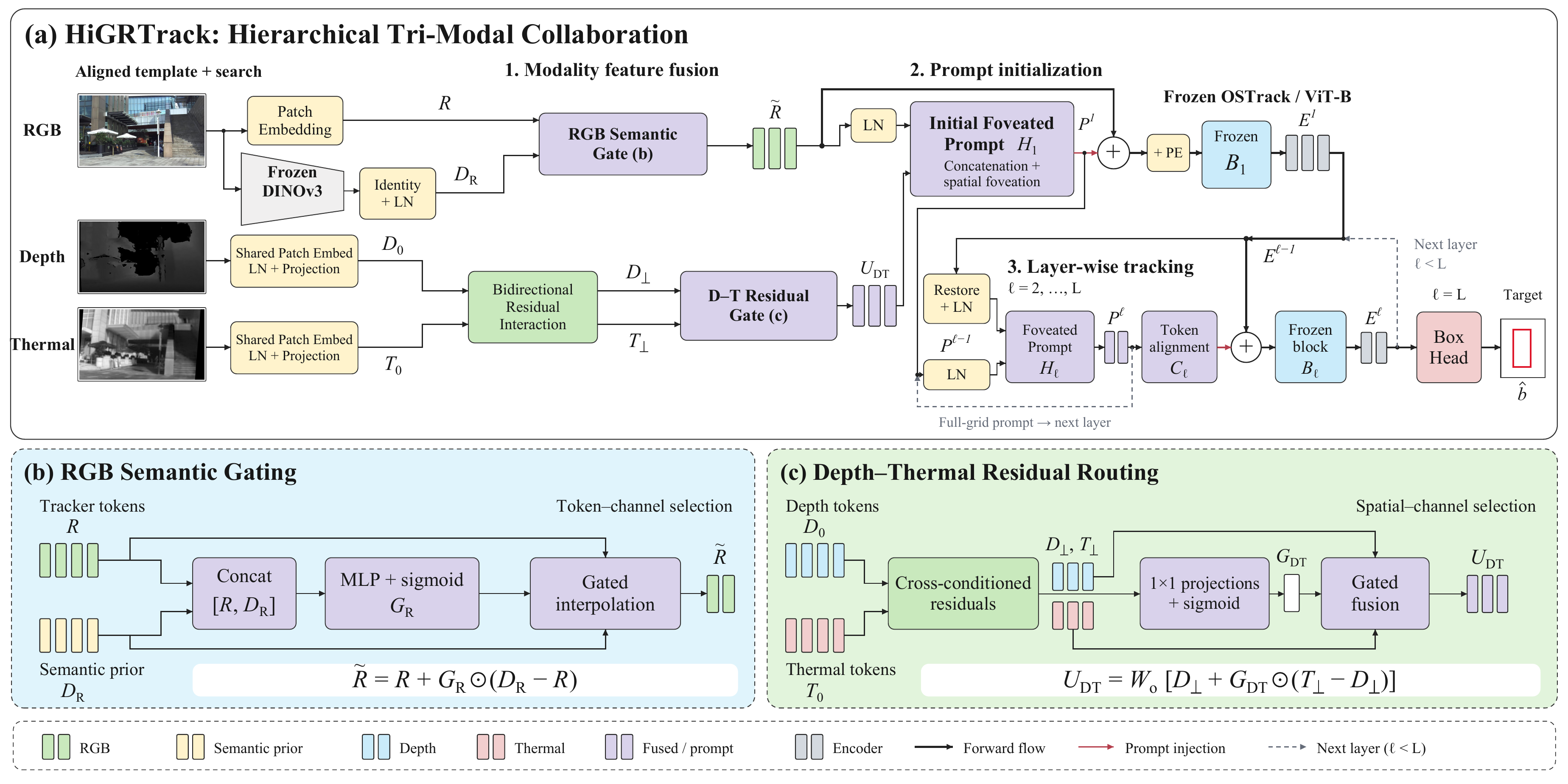}
    \caption{Overview of HiGRTrack. DINOv3 encodes template and search crops
    separately; its ViT-B/16 projection is identity. Gated RGB features feed
    both the prompt interface and backbone. Depth and TIR share patch
    embedding, followed by LN, separate projections, residual interaction,
    and gated fusion. The initial prompt is added to enhanced RGB before
    positional embeddings (PE). The main pipeline shows initialization and recurrent prompts
    $H_\ell$ for layers $\ell=2,\ldots,L$. Dashed arrows pass states to the next
    layer; (b,c) detail RGB semantic gating and depth--thermal residual routing.}
    \label{fig:architecture}
\end{figure*}

\section{Related Work}

\textbf{Multimodal Tracking and Fusion.} RGB-D/T benchmarks support
tracking with auxiliary sensors~\cite{yan2021depthtrack,zhu2023rgbd1k,
li2019rgbt,li2021lasher,zhang2022vtuav}.
Tang et al.~\cite{tang2025omni} organize multimodal visual object tracking
around modality alignment, architectures, and evaluation. Brenner et
al.~\cite{brenner2023fusion} review RGB-D--thermal fusion, including
registration and modality weighting. These perspectives motivate selective
use of aligned complementary evidence in RGB-D-T tracking.

\textbf{Prompt and Temporal Adaptation.} Frozen-backbone adaptation
~\cite{jia2022vpt,chen2022adaptformer,oquab2024dinov2,simeoni2025dinov3}
supports multimodal trackers including ViPT, SDSTrack, UnTrack, and BAT
~\cite{zhu2023vipt,hou2024sdstrack,wu2024untrack,cao2024bat}.
Xu et al.~\cite{xu2026frequency} combine spatial/channel/frequency-aware
visual prompts with multilevel temporal memory. STIFormer
~\cite{xu2026stiformer} propagates temporal tokens across frames and uses
token-guided mixed attention to handle changing modality dominance.
Our $H_\ell$ in Eq.~\ref{eq:prompt} instead recurs over encoder layers for
the current template--search input; it introduces no cross-frame memory bank.

\textbf{Spatial/Channel and Token Interaction.} TBSI
~\cite{hui2023tbsi} uses template-mediated search interaction. CSTNet
~\cite{li2024cstnet} combines channel enhancement, multiscale spatial modeling,
and cross-attention. Equation~\ref{eq:dtgate} uses a pointwise gate on
compact depth--thermal residuals, avoiding an additional all-pairs spatial
attention matrix. MRTTrack~\cite{lai2026mrttrack} selects representative
tokens for masked interaction and constrains non-representative cross-modal
tokens. Equation~\ref{eq:orth} instead forms cross-conditioned residuals
at aligned positions, without a token-selection loss.

\textbf{RGB-D-T Tracking.} RGBDT500 establishes aligned RGB, depth, and
thermal sequences and the RDTTrack baseline~\cite{zhu2025rgbdt500}.
We retain its residual construction, recursive prompts, and spatial foveation
for a frozen RGB tracker. HiGRTrack adds gated DINOv3 RGB semantics and
input-dependent depth--thermal residual routing before prompt injection.

\begin{table*}[!t]
\caption{Comparison on RGBDT500 under one pass evaluation. Higher is better;
bold values indicate the best results among the compared configurations.}
\label{tab:main}
\centering
\begin{tabular}{llcclc}
\toprule
Tracker & Input modality & AUC & DP & Publication & Year \\
\midrule
ATOM~\cite{danelljan2019atom} & RGB & 0.649 & 0.695 & CVPR & 2019 \\
DiMP~\cite{bhat2019dimp} & RGB & 0.662 & 0.710 & ICCV & 2019 \\
PrDiMP~\cite{danelljan2020prdimp} & RGB & 0.676 & 0.698 & CVPR & 2020 \\
ToMP~\cite{mayer2022tomp} & RGB & 0.710 & 0.747 & CVPR & 2022 \\
OSTrack~\cite{ye2022one} & RGB & 0.694 & 0.736 & ECCV & 2022 \\
SeqTrack~\cite{chen2023seqtrack} & RGB & 0.719 & 0.778 & CVPR & 2023 \\
MixFormer~\cite{cui2024mixformer} & RGB & 0.732 & 0.781 & TPAMI & 2024 \\
\midrule
DeT\_ATOM~\cite{yan2021depthtrack} & RGB-D & 0.639 & 0.665 & ICCV & 2021 \\
DeT\_DiMP~\cite{yan2021depthtrack} & RGB-D & 0.667 & 0.700 & ICCV & 2021 \\
SPT~\cite{zhu2023rgbd1k} & RGB-D & 0.706 & 0.757 & AAAI & 2023 \\
ViPT~\cite{zhu2023vipt} & RGB-D & 0.720 & 0.759 & CVPR & 2023 \\
SDSTrack~\cite{hou2024sdstrack} & RGB-D & 0.718 & 0.763 & CVPR & 2024 \\
UnTrack~\cite{wu2024untrack} & RGB-D & 0.733 & 0.776 & CVPR & 2024 \\
\midrule
TBSI~\cite{hui2023tbsi} & RGB-T & 0.690 & 0.749 & CVPR & 2023 \\
ViPT~\cite{zhu2023vipt} & RGB-T & 0.693 & 0.752 & CVPR & 2023 \\
SDSTrack~\cite{hou2024sdstrack} & RGB-T & 0.666 & 0.708 & CVPR & 2024 \\
BAT~\cite{cao2024bat} & RGB-T & 0.713 & 0.782 & AAAI & 2024 \\
UnTrack~\cite{wu2024untrack} & RGB-T & 0.732 & 0.790 & CVPR & 2024 \\
DCEvo+OSTrack~\cite{liu2025dcevo} & RGB-T & 0.706 & 0.741 & CVPR & 2025 \\
\midrule
RDTTrack~\cite{zhu2025rgbdt500} & RGB-D-T & 0.752 & 0.792 & NeurIPS & 2025 \\
\textbf{HiGRTrack} & RGB-D-T & \textbf{0.759} & \textbf{0.804} & \textbf{Proposed} & -- \\
\bottomrule
\end{tabular}
\end{table*}

\section{Proposed Method}

\subsection{Overall Architecture}

Figure~\ref{fig:architecture} shows the two gating decisions.
Let $X_R$, $X_D$, and $X_T$ denote the concatenated template and search patches
for RGB, depth, and thermal inputs. Modality embeddings represent sensor-specific statistics. The RGB branch supplies the main tracking tokens, while depth and
thermal form an auxiliary prompt path. \framework{} calibrates RGB semantics
and routes auxiliary residuals in parallel; the two streams meet before
recurrent prompt injection and center-based box prediction.

The frozen one-stream encoder performs joint attention between the template and
search tokens after each prompt injection. Partitioning queries and keys by
template ($z$) and search ($x$) gives
\begin{equation}
A=\operatorname{softmax}_{\mathrm{row}}\!\left(\frac{QK^\top}{\sqrt{d_k}}\right)=
\begin{bmatrix}A_{zz}&A_{zx}\\A_{xz}&A_{xx}\end{bmatrix},
\label{eq:crossregion}
\end{equation}
where $Q,K\in\mathbb{R}^{N_a\times d_k}$ are the per-head query and key
matrices, $d_k$ is the head dimension, and $N_a$ counts the active tokens.
Softmax normalizes each query row over all key positions.
$A_{zx}$ and $A_{xz}$ exchange information between template and search
regions; prompt injection preserves this one-stream attention structure.

\subsection{RGB Semantic Gating}

DINOv3 produces strong dense features without task-specific labels
\cite{simeoni2025dinov3}. We align its frozen patch tokens with the tracker
tokens and learn interpolation weights from the input features:
\begin{align}
R           &=P_R(X_R),\nonumber\\
D_R         &=\operatorname{LN}\!\left(
               W_p\Phi_{\mathrm{DINOv3}}(X_R)\right),
\label{eq:rgbfeat}\\
G_R         &=\sigma\!\left(W_2\,\operatorname{GELU}
               \left(W_1[R\Vert D_R]\right)\right),\nonumber\\
\widetilde R&=R+G_R\odot(D_R-R).
\label{eq:rgbgate}
\end{align}
Here $N=N_z+N_x$ counts the full template and search grids, $C=768$ is
the backbone width, $\Vert$ concatenates channels, and $\sigma$ is sigmoid.
$\operatorname{LN}$ normalizes channels independently per token.
$G_R\in[0,1]^{N\times C}$ is predicted for every patch and channel. Small values retain tracker detail, while large values incorporate the
semantic prior. Bilinear token-grid alignment is used if the patch layouts
differ; the configured ViT-B/16 branches both use stride 16. This gate blends
two representations of the same RGB input. Gate values are learned mixing
coefficients; their magnitude is not a calibrated measure of sensor reliability.

\subsection{Residual Complementarity Routing}

Depth and thermal embeddings are first projected to a compact hidden space:
$D_0=W_D P_D(X_D)$ and $T_0=W_T P_T(X_T)$. We then compute a cross-conditioned response and subtract a learned
amount of it, retaining the residual construction of RDTTrack. The
resulting depth and thermal residuals are
\begin{equation}
\begin{aligned}
\Pi_{D|T}
  &=\frac{D_0\odot T_0}{\lVert T_0\rVert_2+\epsilon}\odot T_0,\\
D_{\perp}
  &=D_0-\sigma(a)\Pi_{D|T},\\[1pt]
\Pi_{T|D}
  &=\frac{T_0\odot D_0}{\lVert D_0\rVert_2+\epsilon}\odot D_0,\\
T_{\perp}
  &=T_0-\sigma(b)\Pi_{T|D}.
\end{aligned}
\label{eq:orth}
\end{equation}
Here $D_0,T_0\in\mathbb{R}^{N\times h}$ with $h=8$. All products are
elementwise; explicitly, $\lVert T_{0,n,:}\rVert_2=
\sqrt{\sum_{c=1}^{h}T_{0,nc}^2}$ for token $n$, and likewise for $D_0$.
The denominator is broadcast over channels, with $\epsilon=10^{-6}$.
Following the retained RDTTrack block, $a,b$ are learned scalar logits
initialized to $0.5$ and shared across tokens and channels. The subscript
$\perp$ denotes soft cross-conditioned residuals. Our added router blends
them at each position and channel:
\begin{align}
G_{DT}&=\sigma\!\left(C_D(D_{\perp})+C_T(T_{\perp})\right),\nonumber\\
U_{DT}&=W_o\!\left(D_{\perp}+G_{DT}\odot(T_{\perp}-D_{\perp})\right).
\label{eq:dtgate}
\end{align}
$C_D,C_T$ are separate $1\times1$ projections with spatially shared weights,
each mapping $8\to8$ channels. Summing their outputs and applying sigmoid
yields $G_{DT}\in(0,1)^{N\times8}$; $W_o$ maps the blend $8\to768$.
Both gates in Eqs.~\ref{eq:rgbgate} and~\ref{eq:dtgate} implement
$f_i=(1-g_i)x_i+g_i y_i$, so
\[
\min(x_i,y_i)\le f_i\le\max(x_i,y_i),\qquad 0\le g_i\le1.
\]
This coordinatewise bound applies before output projection; it explains
bounded mixing, not statistical confidence or optimization stability.
For $N=64+256$ patch tokens and $h=8$, the two gate projections require
$2Nh^2=40{,}960$ multiply--accumulates, excluding biases and $W_o$.
Their $O(Nh^2)$ cost avoids an additional $O(N^2h)$ all-pairs attention term;
it excludes DINOv3 feature extraction and total tracker runtime.

\subsection{Foveated Deep Prompt Collaboration}

Following the recurrent prompt structure of RDTTrack, the calibrated RGB
map and routed auxiliary representation initialize the first prompt. At later stages, the current encoder representation and the
previous prompt are combined to form the next prompt, which is then injected
into the frozen encoder:
\begin{align}
P^1&=H_1([\widetilde R\Vert U_{DT}]),\nonumber\\
P^\ell&=H_\ell\!\left[\operatorname{LN}(E^{\ell-1})\Vert
                 \operatorname{LN}(P^{\ell-1})\right],
                 \quad \ell>1,\nonumber\\
E^\ell&=B_\ell(E^{\ell-1}+\mathcal{C}_\ell(P^\ell)).
\label{eq:prompt}
\end{align}
$E^0=\widetilde R+\mathrm{PE}$ denotes the enhanced RGB backbone tokens
including positional embeddings.
$H_\ell$ uses $1\times1$ projections and softmax over spatial positions
within each channel, separately for template and search maps. Before
$H_\ell$, pruned encoder tokens are restored to the full grid with zeros;
$\mathcal{C}_\ell$ gathers the search indices retained at the input to block
$\ell$, preserving template order. Thus each prompt and encoder token
correspond to the same location during injection. Here $\ell$ indexes network
depth, not video time: prompts are recomputed for each tracking input.
In the retained foveation operator, spatial softmax weights multiply the
projected RGB/encoder stream before adding the projected auxiliary/previous
prompt stream and projecting back to $C$ channels.

\begin{figure*}[!t]
    \centering
    \begin{minipage}[t]{0.30\textwidth}
        \centering
        \includegraphics[width=\linewidth]{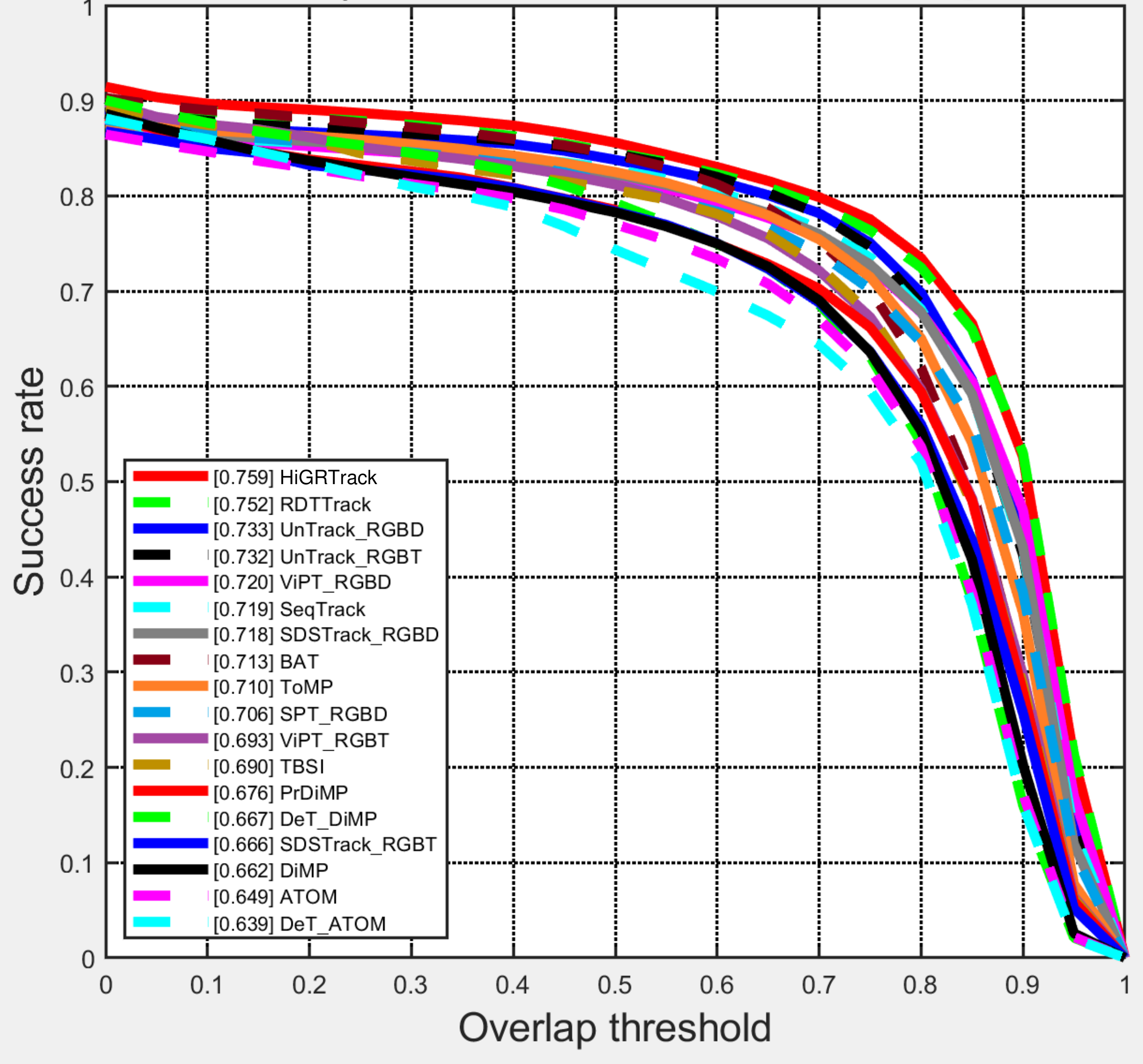}
    \end{minipage}
    \hspace{0.025\textwidth}
    \begin{minipage}[t]{0.30\textwidth}
        \centering
        \includegraphics[width=\linewidth]{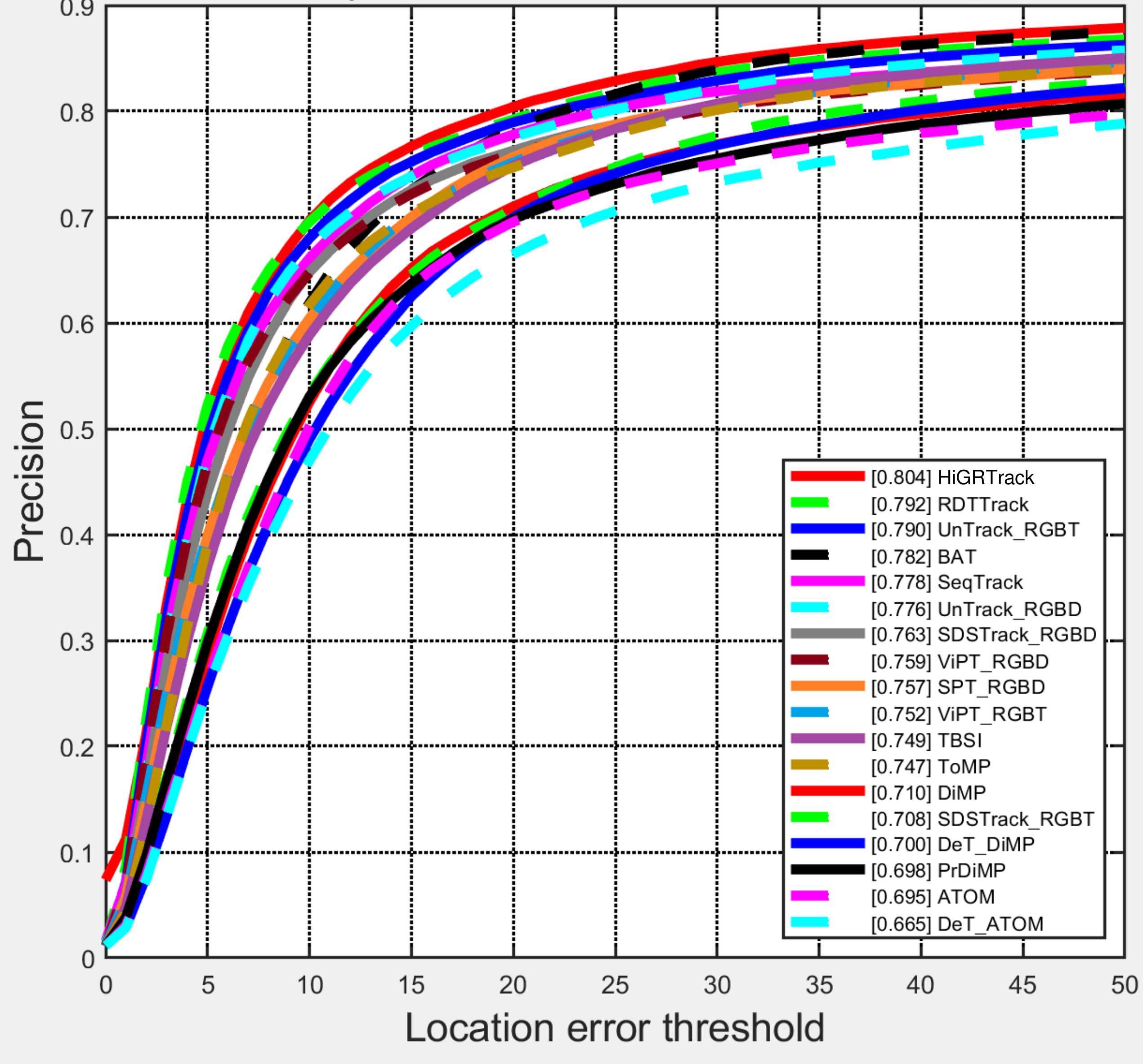}
    \end{minipage}
    \caption{Success (left) and precision (right) curves on RGBDT500.
    HiGRTrack achieves 0.759 AUC and 0.804 DP.}
    \label{fig:supplied-curves}
\end{figure*}

\section{Experiments}

\subsection{Dataset, Protocol, and Metrics}

RGBDT500 contains 400 training and 100 test sequences spanning 66 categories;
the test set has 43.7K synchronized RGB-D-T triplets~\cite{zhu2025rgbdt500}.
Under one-pass evaluation~\cite{wu2015object}, we report success AUC
(area under the IoU-threshold success curve) and DP@20
(fraction of frames with center error within 20 pixels).

\subsection{Implementation Details}

We build on OSTrack with a ViT-B/16 backbone initialized from its epoch 300
checkpoint. A frozen DINOv3 ViT-B/16 provides dense RGB features, while depth
and thermal features are projected to eight channels for auxiliary routing.
Both encoders remain frozen, and only the semantic calibration, auxiliary
routing, and prompt modules are optimized, resulting in 2.04M trainable
parameters, excluding frozen weights. DINOv3 still adds feature-extraction
computation and memory, so this count does not measure inference cost.

All modalities are represented using three input channels. Depth maps are
normalized and converted to JET maps. During training, values above
$\min(3\operatorname{median}(D),10000)$ are clipped. Template and search crops
are resized to 128 and 256 pixels with area factors of 2.0 and 4.0. Frame pairs
are sampled with a maximum interval of 200, producing 60,000 samples per epoch.
Standard spatial and photometric augmentations are applied during training.

Candidate elimination uses configuration indices $[3,6,9]$ (zero-based
in the public RDTTrack implementation). The keep ratio remains
1.0 for the first four epochs and gradually decreases to 0.7 by epoch 20. We
train the model for 60 epochs using AdamW with an initial learning rate of
$4\times10^{-5}$, weight decay of $10^{-4}$, and a global batch size of 64.
The learning rate is reduced by a factor of 10 at epoch 48. The training
objective is $2L_{\mathrm{GIoU}}+5L_1+L_{\mathrm{focal}}$
~\cite{lin2017focal,rezatofighi2019giou}. Inference follows OSTrack using a
Hanning response window and averaged box predictions.

\begin{table}[t]
\caption{Separate ablation configurations on RGBDT500. PW: $1\times1$
D--T gate; DW: $3\times3$ depthwise gate; CA: cross attention;
MD: modality dropout. The last row is Full (HiGRTrack).}
\label{tab:ablation}
\centering
\setlength{\tabcolsep}{3pt}
\setlength{\doublerulesep}{1.5pt}
\renewcommand{\arraystretch}{1.08}
\begin{tabular}{@{}cccccc||cc@{}}
\toprule[0.8pt]
DINOv3 & DINOv2 & PW & DW & CA & MD & DP@20 & AUC \\
\midrule[0.4pt]
           &            & $\checkmark$ &              &              &              & 0.784 & 0.736 \\
           & $\checkmark$ & $\checkmark$ &              &              &              & 0.777 & 0.730 \\
$\checkmark$ &            &              & $\checkmark$ &              &              & 0.787 & 0.744 \\
$\checkmark$ &            & $\checkmark$ &              & $\checkmark$ &              & 0.784 & 0.737 \\
$\checkmark$ &            & $\checkmark$ &              &              & $\checkmark$ & 0.763 & 0.719 \\
$\checkmark$ &            & $\checkmark$ &              &              &              & \textbf{0.804} & \textbf{0.759} \\
\bottomrule[0.8pt]
\end{tabular}
\end{table}

\subsection{Comparison with the State of the Art}

Table~\ref{tab:main} compares 20 RGB, RGB-D, RGB-T, and RGB-D-T baselines.
HiGRTrack achieves the highest AUC/DP of 0.759/0.804, exceeding RDTTrack
by 0.007/0.012. It also improves over UnTrack-RGBD by 0.026 AUC and
UnTrack-RGBT by 0.014 DP. Figure~\ref{fig:supplied-curves} shows the
success and precision curves.

\subsection{Ablation Study}

Full reaches 0.759 AUC/0.804 DP in Table~\ref{tab:ablation}.
The depthwise-gate substitution retains DINOv3 but reduces AUC/DP by
0.015/0.017. This comparison supports the selected fusion operator within
the semantic-enhanced configuration. Cross attention, DINOv2, and modality
dropout yield 0.737/0.784, 0.730/0.777, and 0.719/0.763 AUC/DP,
respectively. Thus encoder presence alone does not reproduce Full's accuracy.

No DINOv3 reaches 0.736 AUC/0.784 DP, below RDTTrack by 0.016/0.008.
This ablation removes semantic enrichment from HiGRTrack; it does not revert
Eq.~\ref{eq:dtgate} to RDTTrack's projection of concatenated residuals
~\cite{zhu2025rgbdt500}. The resulting prompt input therefore differs from
the baseline. Full versus No DINOv3 measures a 0.023/0.020 AUC/DP
removal effect within this design, while Full versus RDTTrack measures the
net 0.007/0.012 improvement between complete trackers. These comparisons
support the complete design without establishing independent or additive
benefits for every component.

These configuration-level results and the local mixing bound do not
establish repeated-seed stability, end-to-end efficiency, or gate calibration.

\section{Conclusion}

We present a hierarchical RGB-D-T tracking framework that calibrates RGB
semantics and routes complementary depth and thermal evidence through deep
prompts. It achieves 0.759 AUC and 0.804 DP on RGBDT500. The ablations support the combined design and selected
fusion operator.

\clearpage
\bibliographystyle{IEEEbib}
\bibliography{references}

\end{document}